\documentclass[letterpaper, 10 pt, conference]{ieeeconf}

\IEEEoverridecommandlockouts
\usepackage{amsmath}
\usepackage{amssymb}
\usepackage{graphicx}
\usepackage{booktabs}

\newtheorem{theorem}{Theorem}
\newtheorem{corollary}{Corollary}

\title{\LARGE \bf
Explicit Bounds on the Hausdorff Distance for Truncated mRPI Sets via Norm-Dependent Contraction Rates
}

\author{Jiaxun Sun$^{1}$, Hengyu Xue$^{2}$, Yuyang Zhao$^{3}$%
\thanks{$^{1}$Jiaxun Sun is with the Department of Mechanical and Process Engineering, ETH Z\"urich, Z\"urich 8092, Switzerland (\texttt{\small jiaxsun@ethz.ch}).}%
\thanks{$^{2}$Hengyu Xue and $^{3}$Yuyang Zhao are with the School of Transportation Science and Engineering, Beihang University, Beijing 100191, China (\texttt{\small hyxue@buaa.edu.cn}, \texttt{\small yuyzhao@buaa.edu.cn}).}%
}

\begin{document}

\maketitle
\thispagestyle{empty}
\pagestyle{empty}

\begin{abstract}
We derive a computable closed-form upper bound on the Hausdorff distance
between a truncated minimal robust positively invariant (mRPI) set and its
infinite-horizon limit. The bound depends only on a disturbance-set size
measure and an induced-norm contraction factor of the system matrix, and it
yields an explicit, fully analytic horizon-selection rule that guarantees a
prescribed approximation tolerance without iterative set computations. The
choice of vector norm enters as a design lever: norm shaping, through
diagonal or Lyapunov-based weighting, tightens both the contraction factor
and the resulting certificate, with direct consequences for robust
invariant-set approximation and tube-based model predictive control (MPC)
constraint tightening. Numerical examples illustrate the accuracy,
scalability, and practical impact of the proposed bound.
\end{abstract}

\section{Introduction}

Robust positively invariant (RPI) sets are fundamental to the analysis and
synthesis of constrained robust Model Predictive Control (MPC)
schemes~\cite{c1,c2,c3}. The minimal RPI (mRPI) set in particular
characterises the steady-state effect of bounded disturbances on linear
systems and underpins constraint tightening in tube-based
MPC~\cite{c6,c13,c15}. Exact computation of the mRPI is generally
intractable, motivating tractable approximations.

A classical and widely used construction represents the mRPI as the infinite
Minkowski series $\mathcal{E}_\infty=\bigoplus_{i=0}^{\infty}A^{i}\mathcal{W}$,
whose truncation $\mathcal{E}_N=\bigoplus_{i=0}^{N-1}A^{i}\mathcal{W}$ yields
a practical \emph{inner} approximation when $A$ is Schur stable and
$\mathcal{W}$ is bounded with $0\in\mathcal{W}$, since
$\mathcal{E}_N\subseteq\mathcal{E}_\infty$ for all~$N$. The
Kolmanovsky--Gilbert construction~\cite{c6} and the refinements of
Rakovi\'c~\textit{et~al.}~\cite{c4,c5}, together with the contraction-based
analyses of~\cite{c12,c14}, compute such approximations through truncated
Minkowski sums, fixed-point iterations, or polytopic relaxations, and
establish $\mathcal{E}_N\to\mathcal{E}_\infty$ under stability and
boundedness assumptions. Subsequent works on optimisation-based invariant
sets and reachable-set aggregations~\cite{c8,c16,c19} enrich the toolbox
and improve scalability in specific settings. However, in many practical
workflows the designer still lacks an \emph{explicit, computable}
truncation-error bound---a closed-form bound on the Hausdorff distance
$d_H(\mathcal{E}_N,\mathcal{E}_\infty)$ that can be evaluated \textit{a
priori} for a prescribed horizon $N$ and a user-selected norm.

This gap matters in tube-based MPC~\cite{c13,c15,c19}, where constraint
tightening depends on outer approximations of the disturbance-induced error
set. Without an explicit characterisation of the truncation error one
resorts to conservative horizons or empirical tuning, unnecessarily
shrinking the feasible region.

Motivated by these considerations, this paper derives a simple analytic
truncation-error certificate for the mRPI Minkowski series and exploits it
for horizon selection. The contributions are:
\begin{itemize}
  \item A closed-form upper bound on $d_H(\mathcal{E}_N,\mathcal{E}_\infty)$
        in terms of an induced-norm contraction factor $\gamma<1$ of $A$ and
        a disturbance-set size measure $r_{\mathcal{W}}$.
  \item An explicit horizon-selection rule guaranteeing
        $d_H(\mathcal{E}_N,\mathcal{E}_\infty)\le\varepsilon$ without
        iterative set computations.
  \item Concrete recipes (diagonal scaling, Lyapunov shaping) for choosing
        the norm so as to tighten the certificate, together with a discussion
        of the joint dependence of $\gamma$ and $r_{\mathcal{W}}$ on that
        choice.
\end{itemize}
Numerical experiments illustrate the tightness, scalability, and tube-MPC
impact of the resulting guarantees.

\section{Preliminaries}

\subsection{System and set operations}
We consider the discrete-time linear system
\begin{equation}
    x_{k+1} = A x_k + w_k, \qquad w_k \in \mathcal{W},
    \label{eq:sys}
\end{equation}
where $A\in\mathbb{R}^{n\times n}$ is Schur stable and
$\mathcal{W}\subset\mathbb{R}^n$ is a compact, convex disturbance set with
$0\in\mathcal{W}$, ensuring that the truncated Minkowski sums introduced
below are monotone. For sets $\mathcal{A},\mathcal{B}\subset\mathbb{R}^n$,
the Minkowski sum is
$\mathcal{A}\oplus\mathcal{B}=\{a+b:a\in\mathcal{A},\,b\in\mathcal{B}\}$,
and the Pontryagin difference is
$\mathcal{X}\ominus\mathcal{Z}=\{x\in\mathbb{R}^n:x+z\in\mathcal{X},\,\forall z\in\mathcal{Z}\}$.

\subsection{Norms, support functions, radius, and the Hausdorff distance}
Let $\|\cdot\|$ be a vector norm on $\mathbb{R}^n$, with \emph{dual norm}
$\|u\|_*:=\sup_{\|x\|\le 1}u^\top x$. For a compact set
$\mathcal{S}\subset\mathbb{R}^n$ we use the \emph{support function}
\begin{equation}
    h_{\mathcal{S}}(u) := \sup_{s\in\mathcal{S}} u^\top s,
    \label{eq:support}
\end{equation}
and the \emph{radius} (in the chosen norm)
$\mathrm{rad}(\mathcal{S}):=\sup_{s\in\mathcal{S}}\|s\|$. The Hausdorff
distance induced by $\|\cdot\|$ is, for compact $\mathcal{A},\mathcal{B}$,
\begin{equation}
d_H(\mathcal{A},\mathcal{B})
=\max\!\Big\{
\sup_{a\in\mathcal{A}}\inf_{b\in\mathcal{B}}\|a-b\|,\;
\sup_{b\in\mathcal{B}}\inf_{a\in\mathcal{A}}\|b-a\|
\Big\}.
\label{eq:hd}
\end{equation}
For compact convex sets it admits the support-function representation
$d_H(\mathcal{A},\mathcal{B})=\sup_{\|u\|_*=1}|h_{\mathcal{A}}(u)-h_{\mathcal{B}}(u)|$~\cite{c8,c14}.

\subsection{Truncated mRPI sets and norm-induced contraction}\label{subsec:mRPI}
The mRPI set associated with $(A,\mathcal{W})$ is
\begin{equation}
    \mathcal{E}_\infty := \bigoplus_{i=0}^{\infty} A^{i}\mathcal{W},
    \qquad
    \mathcal{E}_N := \bigoplus_{i=0}^{N-1} A^{i}\mathcal{W},
    \label{eq:einf-en}
\end{equation}
both compact under Schur stability of $A$ and boundedness of
$\mathcal{W}$. Since $0\in\mathcal{W}$,
$\mathcal{E}_N\subseteq\mathcal{E}_\infty$ for every $N$, so
$\mathcal{E}_N$ is an \emph{inner} approximation. The corresponding tail
set is
\begin{equation}
    \mathcal{T}_N := \bigoplus_{i=N}^{\infty} A^{i}\mathcal{W},
    \qquad
    \mathcal{E}_\infty=\mathcal{E}_N\oplus\mathcal{T}_N.
    \label{eq:tail}
\end{equation}

For any Schur stable $A$ there \emph{exists} a vector norm under which the
induced matrix norm satisfies $\|A\|<1$ (e.g.\ a Lyapunov-shaped norm,
cf.~Sec.~\ref{subsec:norm_opt}); this is in general \emph{not} true for
every norm~\cite{c12,c14}. Throughout the paper we assume such a norm has
been selected and write
\begin{equation}
    \gamma := \|A\| < 1,\qquad
    r_{\mathcal{W}} := \max_{w\in\mathcal{W}} \|w\|.
    \label{eq:gamma-rw}
\end{equation}
Then $\|A^{i}\|\le\gamma^{i}$ and $\|A^{i}w\|\le r_{\mathcal{W}}\gamma^{i}$
for all $i\ge 0$, $w\in\mathcal{W}$, which is the key geometric-decay
property used in the sequel.

\section{Main Results}\label{sec:main_results}

\subsection{Explicit Hausdorff-distance bound}

\begin{theorem}\label{thm:main}
Under the standing assumptions of Sec.~\ref{subsec:mRPI}, for all
$N\in\mathbb{N}$,
\begin{equation}
d_H(\mathcal{E}_N,\mathcal{E}_\infty)
\;\le\;
\frac{r_{\mathcal{W}}\,\gamma^{N}}{1-\gamma}.
\label{eq:main_bound}
\end{equation}
\end{theorem}

\begin{proof}
Since $\mathcal{E}_N\subseteq\mathcal{E}_\infty$, the one-sided Hausdorff
term $\sup_{x\in\mathcal{E}_N}\inf_{y\in\mathcal{E}_\infty}\|x-y\|$
vanishes, so by~\eqref{eq:hd},
\begin{equation}
d_H(\mathcal{E}_N,\mathcal{E}_\infty)
=\sup_{y\in\mathcal{E}_\infty}\inf_{x\in\mathcal{E}_N}\|y-x\|.
\label{eq:hd-reduce}
\end{equation}
Fix any $y\in\mathcal{E}_\infty$. By~\eqref{eq:tail} there exist
$x\in\mathcal{E}_N$ and $t\in\mathcal{T}_N$ with $y=x+t$; in particular
the residual $t=t(x,y):=y-x$ depends on $(x,y)$. Hence
\begin{equation}
\inf_{x'\in\mathcal{E}_N}\|y-x'\|\;\le\;\|t(x,y)\|
\;\le\;\sup_{t'\in\mathcal{T}_N}\|t'\|\;=\;\mathrm{rad}(\mathcal{T}_N).
\label{eq:inf-bound-t}
\end{equation}
The right-hand side does not depend on $y$. Taking the supremum over
$y\in\mathcal{E}_\infty$ in~\eqref{eq:hd-reduce} and combining with
\eqref{eq:inf-bound-t} gives
\begin{equation}
d_H(\mathcal{E}_N,\mathcal{E}_\infty)\;\le\;\mathrm{rad}(\mathcal{T}_N).
\label{eq:hd-le-rad}
\end{equation}
For the partial tails
$\mathcal{T}_N^{K}:=\bigoplus_{i=N}^{K}A^{i}\mathcal{W}$, the radius is
subadditive under Minkowski sums (because $0\in A^{i}\mathcal{W}$ for all
$i$)~\cite{c1,c14}, so
\begin{equation}
\mathrm{rad}(\mathcal{T}_N^{K})
\le\sum_{i=N}^{K}\mathrm{rad}(A^{i}\mathcal{W})
\le\sum_{i=N}^{K}\|A^{i}\|\,\mathrm{rad}(\mathcal{W})
\le r_{\mathcal{W}}\sum_{i=N}^{K}\gamma^{i}.
\label{eq:rad-tail-finite}
\end{equation}
Letting $K\to\infty$ yields
$\mathrm{rad}(\mathcal{T}_N)\le r_{\mathcal{W}}\gamma^{N}/(1-\gamma)$, which
combined with~\eqref{eq:hd-le-rad} proves~\eqref{eq:main_bound}.
\end{proof}

\subsection{Two equivalent viewpoints on the bound}

The certificate~\eqref{eq:main_bound} admits two complementary derivations
that are useful for different purposes.

\paragraph*{Operator-theoretic interpretation}
Define the affine set-operator
$\Phi(\mathcal{S}):=A\mathcal{S}\oplus\mathcal{W}$. Under any induced norm
with $\|A\|=\gamma<1$, $\Phi$ is a $\gamma$-contraction on the space of
nonempty compact sets equipped with $d_H$~\cite{c1,c14}. Hence
$\mathcal{E}_\infty$ is its unique fixed point, and starting from
$\mathcal{S}_0=\{0\}$ the Picard iterates satisfy
$\Phi^{N}(\{0\})=\mathcal{E}_N$. From this perspective $\mathcal{T}_N$ is
the residual of the contractive series and~\eqref{eq:main_bound} is its
summed magnitude.

\paragraph*{Support-function viewpoint}
Equivalently, using the dual representation of $d_H$,
\begin{equation*}
h_{\mathcal{E}_\infty}(u)-h_{\mathcal{E}_N}(u)
=\sum_{i=N}^{\infty}h_{A^{i}\mathcal{W}}(u),
\qquad
h_{A^{i}\mathcal{W}}(u)=h_{\mathcal{W}}(A^{i\top}u).
\end{equation*}
Since $|h_{\mathcal{W}}(v)|\le r_{\mathcal{W}}\|v\|_*$ and
$\|A^{\top}\|_*=\|A\|=\gamma$ for induced norms,
$|h_{A^{i}\mathcal{W}}(u)|\le r_{\mathcal{W}}\gamma^{i}$ whenever
$\|u\|_*=1$, and the geometric tail is recovered. This view also makes
explicit how the chosen norm influences truncation accuracy through both
$\gamma$ and the dual norm.

\subsection{Minimal truncation index}\label{subsec:Nmin}

\begin{corollary}[Minimal truncation index]\label{cor:Nmin}
Let $\varepsilon>0$ and $c:=\varepsilon(1-\gamma)/r_{\mathcal{W}}$. If
$c\ge 1$ then~\eqref{eq:main_bound} already gives
$d_H(\mathcal{E}_0,\mathcal{E}_\infty)\le\varepsilon$ and one may take
$N_{\min}=0$. Otherwise $c\in(0,1)$ and the condition
$d_H(\mathcal{E}_N,\mathcal{E}_\infty)\le\varepsilon$ is implied by
\begin{equation}
N\;\ge\;\frac{\ln c}{\ln\gamma}
\;=\;\frac{\ln\!\big(\varepsilon(1-\gamma)/r_{\mathcal{W}}\big)}{\ln\gamma},
\label{eq:Nmin}
\end{equation}
so an admissible choice is
$N_{\min}(\varepsilon)=\big\lceil\ln c/\ln\gamma\big\rceil$.
\end{corollary}

\subsection{Improving $\gamma$ by norm choice}\label{subsec:norm_opt}

Because $\gamma=\|A\|$ depends on the chosen norm, the
certificate~\eqref{eq:main_bound} carries explicit design freedom. For any
Schur $A$, $\rho(A)\le\|A\|$ in any induced norm, and norms exist under
which $\|A\|$ approaches $\rho(A)$~\cite{c12,c14}. Two practical and
constructive recipes are:

\textit{(i) Diagonal scaling.}  Take $\|x\|_D:=\|Dx\|_2$ with $D\succ 0$
diagonal, so that $\gamma=\|DAD^{-1}\|_2$. The optimal $D$ minimising
$\|DAD^{-1}\|_2$ can be obtained by a small SDP, but a cheap and
effective heuristic is $D=\mathrm{diag}(\sqrt{P_{ii}})$ from the Lyapunov
solve below; this typically halves the gap between $\|A\|_2$ and
$\rho(A)$ at negligible cost.

\textit{(ii) Lyapunov-shaped quadratic norm.}  Solve the discrete
Lyapunov equation
\begin{equation}
A^{\top}P A - P = -Q,\qquad Q\succ 0,
\label{eq:dlyap}
\end{equation}
for $P\succ 0$. In the norm $\|x\|_P:=\sqrt{x^{\top}P x}$,
$$
\|Ax\|_P^{2} = x^{\top}A^{\top}PA\,x
= x^{\top}P x - x^{\top}Q x
\le \big(1-\tfrac{\lambda_{\min}(Q)}{\lambda_{\max}(P)}\big)\|x\|_P^{2},
$$
giving $\gamma_P^{2}\le 1-\lambda_{\min}(Q)/\lambda_{\max}(P)<1$ by
construction. A single discrete Lyapunov solve with $Q=I$ already provides
a usable contraction factor and underlies our experiments.

\paragraph*{Joint dependence of $\gamma$ and $r_{\mathcal{W}}$}
The bound~\eqref{eq:main_bound} depends on both $\gamma$ and
$r_{\mathcal{W}}$, and the latter also depends on the chosen norm: a heavier
$P$ that shrinks $\gamma$ may simultaneously inflate $r_{\mathcal{W}}$ for a
fixed disturbance set $\mathcal{W}$. The relevant figure of merit, made
explicit by~\eqref{eq:Nmin}, is the bound's intercept at $N=0$,
\begin{equation}
\beta(\|\cdot\|) := \frac{r_{\mathcal{W}}}{1-\gamma},
\label{eq:beta}
\end{equation}
which dominates~$N_{\min}(\varepsilon)$ multiplicatively.
Sec.~\ref{sec:numerical} reports both quantities and shows that Lyapunov
shaping reduces $\beta$ even when $r_{\mathcal{W}}$ grows.

\section{Practical Use in Tube MPC}\label{sec:guidelines}

In tube-based MPC the relevant matrix is the closed-loop matrix
$A_{\mathrm{cl}}=A+BK$ obtained from the local feedback
$u=K x+v$ that pre-stabilises the error dynamics. We assume
$A_{\mathrm{cl}}$ is Schur stable, and apply the results of
Sec.~\ref{sec:main_results} with $A\leftarrow A_{\mathrm{cl}}$.

\subsection{Tube construction and constraint tightening}

Let $\mathbb{B}_r:=\{x\in\mathbb{R}^n:\|x\|\le r\}$. By
Theorem~\ref{thm:main}, $\mathcal{T}_N\subseteq\mathbb{B}_{r_N}$ with
$r_N := r_{\mathcal{W}}\gamma^{N}/(1-\gamma)$, so a certified outer
approximation of $\mathcal{E}_\infty$ that is usable as a tube cross-section
is
\begin{equation}
\mathcal{Z} \;:=\; \mathcal{E}_N \oplus \mathbb{B}_{r_N},\qquad
\mathcal{E}_N=\bigoplus_{i=0}^{N-1} A_{\mathrm{cl}}^{i}\mathcal{W}.
\label{eq:tube}
\end{equation}
For polytopic state constraints $\mathcal{X}=\{x:Hx\le h\}$ with rows
$H_i^{\top}$, the Pontryagin difference admits the row-wise representation
\begin{equation}
\mathcal{X}\ominus\mathcal{Z}
=\big\{x:\,H_i^{\top}x\le h_i - h_{\mathcal{E}_N}(H_i^{\top}) - r_N\|H_i^{\top}\|_*,\;\forall i\big\},
\label{eq:tighten-state}
\end{equation}
since $h_{\mathcal{Z}}(v)=h_{\mathcal{E}_N}(v)+r_N\|v\|_*$. Input
constraints $\mathcal{U}=\{u:Gu\le g\}$ are tightened analogously through
$K\mathcal{Z}$. In~\eqref{eq:tube}--\eqref{eq:tighten-state}, the only
horizon-dependent quantity is the scalar $r_N$, while the chosen norm
enters through $\gamma$ and the dual norm $\|\cdot\|_*$.

\subsection{Choosing the truncation index and the norm}
The recipe is:
\begin{enumerate}
    \item Pick $\|\cdot\|$ following Sec.~\ref{subsec:norm_opt}: the
          Euclidean norm is the cheapest baseline; a Lyapunov-shaped norm
          via~\eqref{eq:dlyap} is the most effective when
          $A_{\mathrm{cl}}$ is anisotropic.
    \item Compute $\gamma=\|A_{\mathrm{cl}}\|$ and
          $r_{\mathcal{W}}=\max_{w\in\mathcal{W}}\|w\|$ in that norm.
    \item For a given tolerance $\varepsilon$, take $N\ge N_{\min}(\varepsilon)$
          from Corollary~\ref{cor:Nmin}; a small margin
          $N=N_{\min}(\varepsilon)+\Delta$ may be added to hedge against
          modelling mismatch or downstream outer approximations of
          $\mathcal{E}_N$.
    \item Build $\mathcal{Z}$ from~\eqref{eq:tube} and tighten
          via~\eqref{eq:tighten-state}.
\end{enumerate}
The certificate is most effective when (i)~$n$ is moderate or large so
that direct mRPI computation is costly, (ii)~$\mathcal{W}$ is box- or
polytope-bounded so $r_{\mathcal{W}}$ is easy to evaluate, and
(iii)~$A_{\mathrm{cl}}$ is anisotropic so norm shaping reduces $\gamma$.

\section{Numerical Examples}\label{sec:numerical}

This section evaluates the bound on increasingly challenging systems and
illustrates its impact on tube MPC. All systems and seeds are stated
explicitly; the corresponding scripts reproduce every figure.

\subsection{Experiment 1: Six-dimensional mRPI}\label{subsec:exp1}

A randomly generated $6$-dimensional Schur stable system with
$\mathcal{W}=[-0.1,0.1]^{6}$ is used. The matrix $A$ is drawn from
$\mathcal{N}(0,I_6)$ with seed~$1$ and rescaled by $1/(1.5\,\|A\|_2)$, so
that $\|A\|_2=1/1.5\approx 0.6667$. We use the Euclidean norm, giving
$\gamma=0.6667$ and $r_{\mathcal{W}}=0.1\sqrt{6}\approx 0.2449$. The
numerical Hausdorff distance
$d_H^{\mathrm{num}}(\mathcal{E}_N,\mathcal{E}_\infty)$ is approximated by
support-function evaluations along $2000$ sampled dual-unit directions, with
$\mathcal{E}_\infty$ replaced by the long truncation $\mathcal{E}_{K_\infty}$,
$K_\infty=200$.

\begin{figure}[t]
    \centering
    \includegraphics[width=0.47\textwidth]{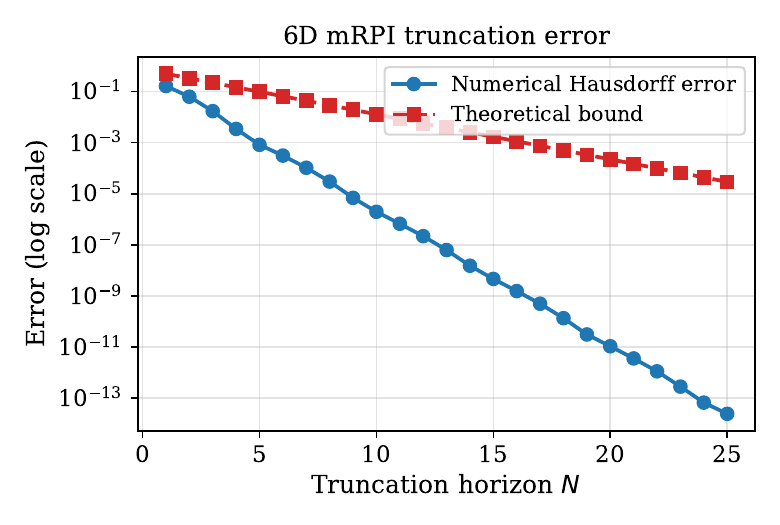}
    \caption{Experiment~1 (6D, Euclidean norm): numerical Hausdorff error
    (blue) and analytic bound~\eqref{eq:main_bound} (red). Both decay
    geometrically with slope governed by $\gamma$.}
    \label{fig:exp1_6d}
\end{figure}

Figure~\ref{fig:exp1_6d} shows that the numerical error and the analytic
bound~\eqref{eq:main_bound} both decay geometrically. The bound is
multiplicatively conservative, but its \emph{slope} on the semilog plot
matches $\gamma^{N}$. This is precisely what drives
$N_{\min}(\varepsilon)$ in~\eqref{eq:Nmin}: a constant proportionality
factor between bound and true error translates into an additive constant in
$N_{\min}$, so the certificate yields the right order even when the
multiplicative gap is several orders of magnitude. In other words, although
the bound becomes increasingly conservative in absolute ratio, it remains
useful as a horizon selector for any prescribed $\varepsilon$.

\subsection{Experiment 2: Norm shaping}\label{subsec:exp2}

We compare three induced norms on a deliberately anisotropic $6$D system
with singular values $(0.85,0.8,0.55,0.4,0.25,0.15)$ (so
$\rho(A)\approx 0.555$ and $\|A\|_2=0.85$):
(a)~the Euclidean norm,
(b)~a diagonal Lyapunov norm $\|x\|_D=\|Dx\|_2$ with
$D=\mathrm{diag}(\sqrt{P_{ii}})$,
(c)~the full Lyapunov-shaped norm $\|x\|_P$, with $P\succ 0$ from
$A^{\top}PA-P=-I$.
The contraction factor $\gamma=\|A\|$ is reduced through norm shaping in
two distinct ways: the diagonal weighting captures only the principal
diagonal of $P$ and is therefore an inexpensive partial step; the full
Lyapunov norm directly measures contraction through the same quadratic
form that certifies stability, and so brings $\gamma$ closer to
$\rho(A)$.

\begin{table}[t]
\centering
\caption{Experiment~2: contraction factor, disturbance radius, and the
intercept $\beta=r_{\mathcal{W}}/(1-\gamma)$ for three induced norms.}
\label{tab:exp2}
\begin{tabular}{lccc}
\toprule
Norm & $\gamma$ & $r_{\mathcal{W}}$ & $\beta=r_{\mathcal{W}}/(1-\gamma)$ \\
\midrule
(a) Euclidean         & 0.850 & 0.245 & 1.633 \\
(b) Diagonal Lyapunov & 0.808 & 0.296 & 1.543 \\
(c) Full Lyapunov     & 0.721 & 0.346 & 1.241 \\
\bottomrule
\end{tabular}
\end{table}

\begin{figure}[t]
    \centering
    \includegraphics[width=0.47\textwidth]{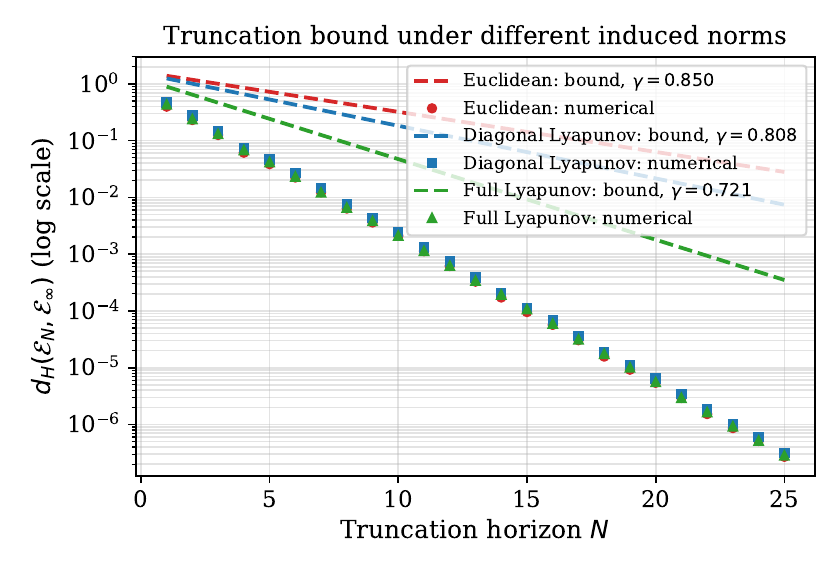}
    \caption{Experiment~2: analytic bound (dashed) and numerical Hausdorff
    error (markers) under three induced norms. Norm shaping shifts the
    analytic curve downward at every $N$ even though $r_{\mathcal{W}}$
    increases.}
    \label{fig:exp2}
\end{figure}

Table~\ref{tab:exp2} reports $(\gamma,r_{\mathcal{W}},\beta)$ for each
norm. Lyapunov shaping reduces $\gamma$ from $0.85$ to $0.72$ but also
increases $r_{\mathcal{W}}$ from $0.25$ to $0.35$. The combined effect is a
\emph{net} reduction of the intercept~$\beta$, hence a strictly tighter
bound for every $N\ge 0$ (Fig.~\ref{fig:exp2}). This concretely
illustrates the joint trade-off discussed in Sec.~\ref{subsec:norm_opt}:
the certificate's quality is governed by $\beta$, not by $\gamma$ alone.

\subsection{Experiment 3: High-dimensional behaviour}\label{subsec:exp3}

We repeat Experiment~1 in dimensions $n\in\{10,15,20\}$ with
$\mathcal{W}=[-0.05,0.05]^{n}$, drawing each $A$ from $\mathcal{N}(0,I_n)$
(seeds $1001$, $1002$, $1003$). To make the contraction factor exactly
comparable across dimensions we normalise each $A$ \emph{by its spectral
norm} so that $\|A\|_2=0.8$, giving $\gamma=0.8$ for every $n$. Note that
this is the correct scaling: normalising by the spectral radius alone does
not control the induced 2-norm of a non-symmetric matrix and would
invalidate~\eqref{eq:main_bound}.

\begin{figure}[t]
    \centering
    \includegraphics[width=0.47\textwidth]{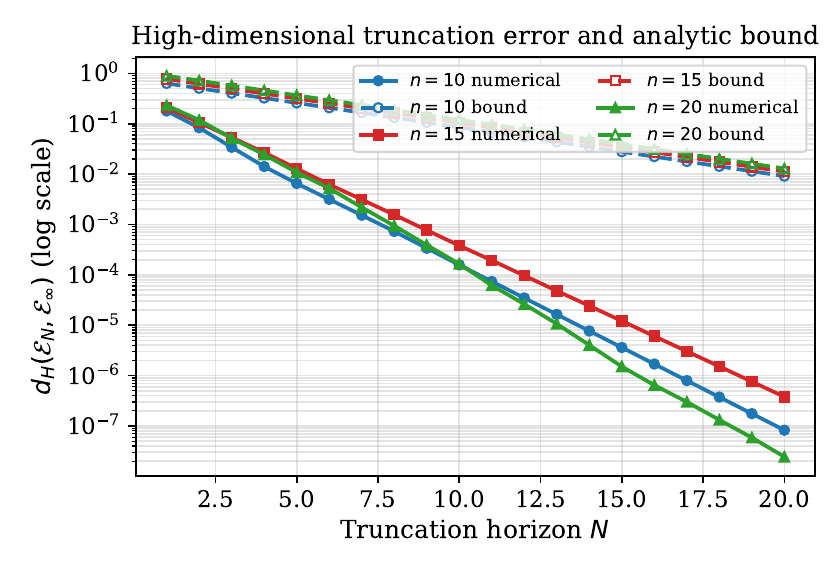}
    \caption{Experiment~3: numerical errors (solid, filled markers) and
    analytic bounds (dashed, hollow markers) for $n=10,15,20$. With
    $\|A\|_2$ normalised to $0.8$, $\gamma=0.8$ for every $n$.}
    \label{fig:exp3_hd}
\end{figure}

Figure~\ref{fig:exp3_hd} plots, for each $n$, the numerical error (solid
line, filled marker) and the analytic bound (dashed line, hollow marker);
distinct colours and marker shapes are used per dimension to avoid the
overlap of the previous version. The numerical errors decay geometrically
and remain below the bound across all dimensions, confirming that the
certificate scales gracefully with $n$. The intercept~$\beta$ scales with
$r_{\mathcal{W}}\propto\sqrt{n}$ in this Euclidean-norm box example,
reflecting the fact that the disturbance set is larger in higher
dimensions; this growth is mild compared with the cost of computing the
exact mRPI in $n=20$.

\subsection{Experiment 4: Tube MPC with the truncation bound}\label{subsec:exp4}

We use the $2$D system
\begin{equation*}
A=\!\begin{bmatrix}0.8 & 0.2\\ 0 & 0.9\end{bmatrix}\!,\;
B=\!\begin{bmatrix}1\\ 0.2\end{bmatrix}\!,\;
K=\!\begin{bmatrix}-0.6 & -0.1\end{bmatrix}\!,
\end{equation*}
state constraints $\mathcal{X}=[-2,2]^{2}$, input constraint $|u|\le 1$,
and $\mathcal{W}=[-0.05,0.05]^{2}$. In the Euclidean norm,
$\gamma=\|A_{\mathrm{cl}}\|_2\approx 0.891$ and
$r_{\mathcal{W}}=\|(0.05,0.05)\|_2\approx 0.0707$, so the coarse
geometric-series bound gives a baseline ball radius
$r_{\mathcal{W}}/(1-\gamma)\approx 0.650$.

Setting $\varepsilon=10^{-2}$, Corollary~\ref{cor:Nmin} yields $N=37$ and
$r_N\approx 9.18\times 10^{-3}$. The proposed certified tube cross-section
is $\mathcal{Z}=\mathcal{E}_N\oplus\mathbb{B}_{r_N}$, computed by
explicit Minkowski summation of the polytope vertices of
$A_{\mathrm{cl}}^{i}\mathcal{W}$. We compare it to the baseline
$\mathcal{Z}_{\mathrm{base}}=\mathbb{B}_{r_{\mathcal{W}}/(1-\gamma)}$.

\begin{figure}[t]
    \centering
    \includegraphics[width=0.42\textwidth]{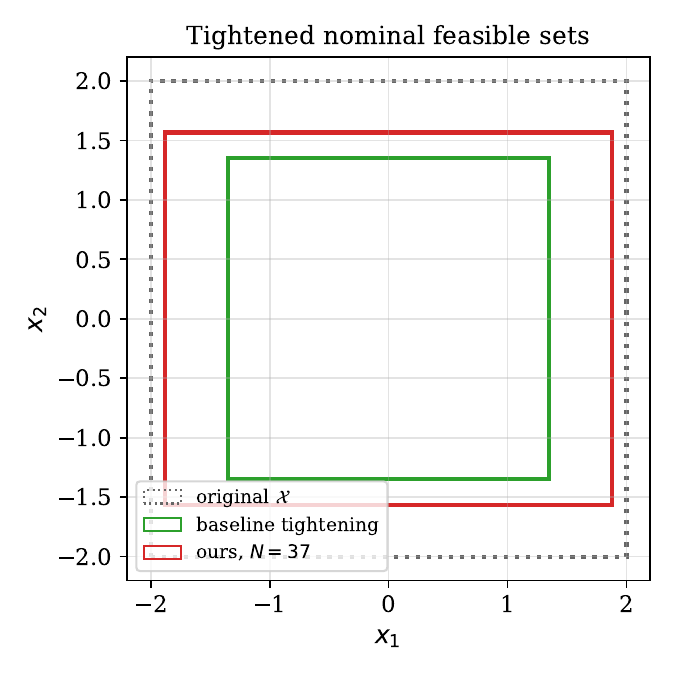}
    \caption{Experiment~4: tightened nominal feasible sets. The baseline
    (green) shrinks $\mathcal{X}$ symmetrically by $0.65$ in every
    direction; the proposed tightening (red) leverages $\mathcal{E}_N$ and
    yields a strictly larger feasible region.}
    \label{fig:exp4_feasible}
\end{figure}

\begin{figure}[t]
    \centering
    \includegraphics[width=0.42\textwidth]{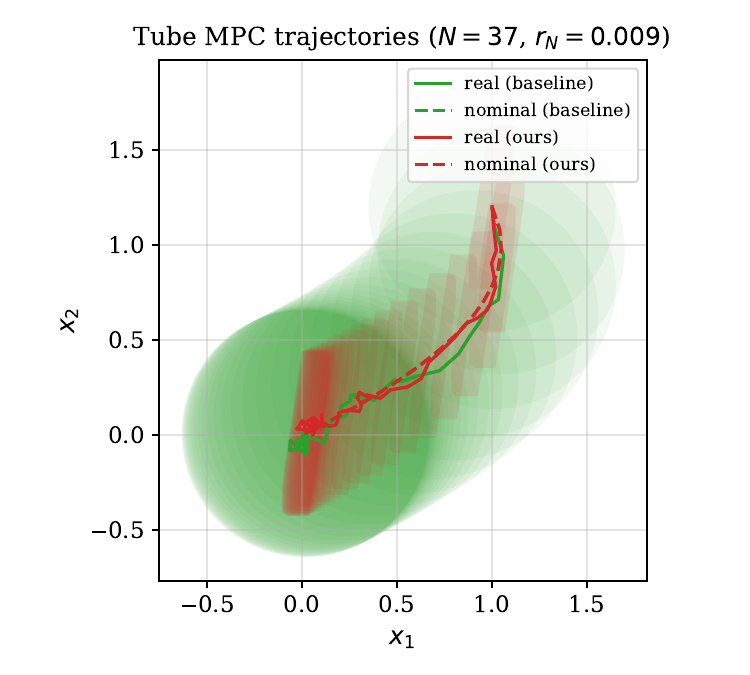}
    \caption{Experiment~4: closed-loop trajectories. Real (solid) and
    nominal (dashed) trajectories for the baseline (green) and the
    proposed method (red), with tubes shown as shaded regions. The
    proposed tube, built from $\mathcal{E}_N$ plus the certified tail
    $\mathbb{B}_{r_N}$, is much smaller while still containing the real
    trajectory.}
    \label{fig:exp4_tube}
\end{figure}

Figure~\ref{fig:exp4_feasible} shows that the proposed tightening gives a
substantially larger nominal feasible region (axis-aligned half-widths
$1.88\times 1.57$ versus $1.35\times 1.35$ for the baseline). Closed-loop
trajectories under random disturbances stay inside the certified tubes
(Fig.~\ref{fig:exp4_tube}), and the proposed tube remains tight while
robustly containing the real trajectory.

\paragraph*{Conservatism in MPC}
Two observations apply when using the bound inside tube MPC. First, the
certified tail radius $r_N$ enters the
tightening~\eqref{eq:tighten-state} only \emph{additively}, so the
multiplicative gap visible in
Figs.~\ref{fig:exp1_6d}--\ref{fig:exp3_hd} is not amplified at the
constraint level: increasing $N$ by a constant offset shrinks $r_N$ by a
constant factor, and the contribution of $r_N$ to the tightening becomes
negligible long before the nominal problem becomes infeasible. Second,
$\beta$ in~\eqref{eq:beta} only enters through the polytope
$\mathcal{E}_N$, whose support function is a finite Minkowski sum of
$N$ tractable terms; thus high-dimensional cases are handled by row-wise
support-function evaluation rather than explicit polytope intersection.

\paragraph*{Sensitivity to $r_{\mathcal{W}}$}
The bound depends on $r_{\mathcal{W}}$ only through an upper estimate. If
the disturbance set is uncertain, replacing $r_{\mathcal{W}}$ by any
over-estimate $r_{\mathcal{W}}^{+}\ge r_{\mathcal{W}}$ produces a valid
(though more conservative) bound. A first-order sensitivity from
\eqref{eq:Nmin}, $\partial N_{\min}/\partial r_{\mathcal{W}}=
-1/(r_{\mathcal{W}}\ln\gamma)>0$, shows that a multiplicative
mis-specification of $r_{\mathcal{W}}$ by a factor $\kappa$ shifts
$N_{\min}$ by $\ln\kappa/|\ln\gamma|$, i.e.\ logarithmically in $\kappa$.
The framework therefore degrades gracefully under imperfect disturbance
characterisation.

\section{Conclusion}

We presented a computable closed-form upper bound on the Hausdorff distance
between a truncated mRPI set and its infinite-horizon limit, expressed as
$r_{\mathcal{W}}\gamma^{N}/(1-\gamma)$ with $\gamma$ an induced-norm
contraction factor. The bound enables explicit horizon selection without
iterative set computations, and the single figure of merit
$\beta=r_{\mathcal{W}}/(1-\gamma)$ governs the certificate's quality. We
gave concrete recipes (diagonal scaling, Lyapunov shaping) for reducing
$\gamma$ and discussed the joint dependence of $\beta$ on the chosen
norm. Numerical experiments validated the geometric decay and scalability
of the bound and demonstrated reduced conservatism and enlarged nominal
feasibility in tube MPC. Future work includes extending the certificate to
parametric and time-varying $A$, and integrating the norm choice into a
joint MPC--invariant-set design.


\end{document}